\documentclass[11pt,letterpaper]{article}
\usepackage{cogsys}
\usepackage[T1]{fontenc}
\usepackage{times}
\usepackage[pdftex]{graphicx} 
\usepackage{algorithm}
\usepackage{algpseudocode}

\usepackage{natbib}
\cogsysheading{Proceedings of the Ninth Goal Reasoning Workshop}{2021}{1-14}{Submitted 10/2021; published 11/2021}{© 2021 Goal Reasoning Workshop. All rights reserved.}

\ShortHeadings{Human-Centric Goal Reasoning with Ripple-Down Rules}
 {K.\ Brameld, G.\ Castro, c.\ Sammut, M.\ Roberts and D.W.\ Aha}

\usepackage{fancyhdr}
\begin{document}

\thispagestyle{fancy}
\cfoot{ \begin{small} \textbf{DISTRIBUTION STATEMENT A.} Approved for public release; distribution is unlimited. \end{small}}

\title{Human-Centric Goal Reasoning with Ripple-Down Rules}
 
\author{Kenji Brameld}{kenjibrameld@gmail.com}
\author{Germ\'an Castro}{german.castro@unsw.edu.au}
\author{Claude Sammut}{c.sammut@unsw.edu.au}
\address{School of Computer Science and Engineering, The University of New South Wales, Australia}
\author{Mark Roberts}{mark.roberts@nrl.navy.mil}
\author{David W. Aha}{david.aha@nrl.navy.mil}
\address{Navy Center for Applied Research in AI, Naval Research Laboratory, Washington, DC 20375 USA}
\vskip 0.2in
 
\begin{abstract}
ActorSim is a goal reasoning framework developed at the Naval Research Laboratory. Originally, all goal reasoning rules were hand-crafted. This work extends ActorSim with the capability of learning by demonstration, that is, when a human trainer disagrees with a decision made by the system, the trainer can take over and show the system the correct decision. The learning component uses Ripple-Down Rules (RDR) to build new decision rules to correctly handle similar cases in the future. The system is demonstrated using the RoboCup Rescue Agent Simulation, which simulates a city-wide disaster, requiring emergency services, including fire, ambulance and police, to be dispatched to different sites to evacuate civilians from dangerous situations. The RDRs are implemented in a scripting language, FrameScript, which is used to mediate between ActorSim and the agent simulator. Using Ripple-Down Rules, ActorSim can scale to an order of magnitude more goals than the previous version.
\end{abstract}

\section{Introduction} 
 
Actor Simulator (ActorSim) is a toolkit for studying situated autonomy \citep{Roberts2016ActorSimAT}. It implements a goal lifecycle that enables an autonomous agent to select and prioritise goals and monitor the progress of plans to achieve those goals. We extend the ActorSim goal reasoning framework \citep{roberts2014iterative, roberts_goal_2021} with machine learning capabilities that enable the system to be taught goal selection and prioritisation strategies by a trainer. ActorSim allows users to plug in simulations of a variety of domains and   different planners. Implemented in Java, it provides a goal reasoning mechanism that is largely independent of the planner and planning domain.

ActorSim incorporates rules for formulating and selecting goals. In the original version, these were coded in Java, the implementation language of ActorSim. Although they were coded in a modular way so that they can be updated and adapted for different domains, there was still are requirement to code directly in Java, making it difficult for someone not familiar with the internals of ActorSim to change the rules. Furthermore, the rules are hand-crafted, meaning that the programmers must anticipate all the goal reasoning strategies that are likely to be employed. The system is more flexible if it is capable of learning the rules by example.

Goal reasoning rules are expressed in a scripting language, FrameScript~\citep{McGill:2019ws}. FrameScript includes a knowledge acquisition system based on Compton’s Ripple-Down Rules (RDR)~\citep{Compton:2021wk}. This is used to implement behavioural cloning~\citep{michie:90, sammut:92}, which is a form of learning by demonstration that aims to reproduce the skill of an expert operator. RDRs were original designed to make it easy for pathologists to teach a knowledge-based system to interpret blood tests. They have since been adapted for a variety of tasks, including teaching an auto-pilot to fly an aircraft in a flight simulator~\citep{shiraz:98}, question answering~\citep{Nguyen:2017vu}, and many more.

The main contributions of this paper are: the application of RDRs to goal reasoning; the first use of RDRs in ActorSim; and the integration of RDRs with a PDDL planner. Preliminary results suggest that RDRs can improve the number of goals ActorSim can manage by at least ten times. 

The remaining sections in this paper describe the goal lifecycle, the rescue domain, in which the system has been evaluated, and the operation of the knowledge acquisition system.

\section{Goal Lifecycle}

\begin{figure}[t]
\begin{center}
\includegraphics[width=8cm]{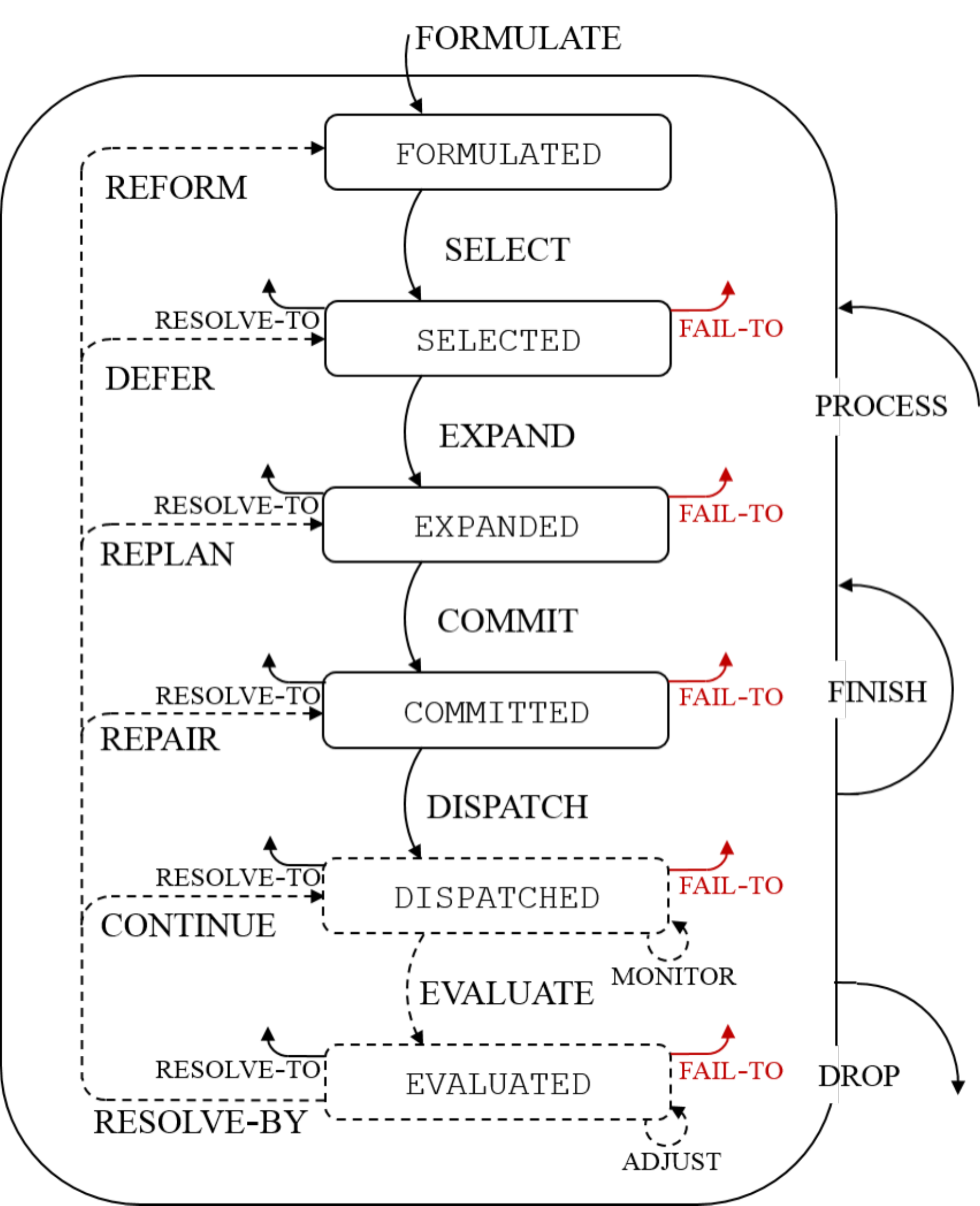}
\caption{ActorSim goal life cycle from \cite{roberts2014iterative}} 
\label{goal_cycle}
\end{center}
\end{figure}

The context for ActorSim is an agent embedded in a dynamic environment. It must formulate its own goals to achieve an overall objective. Because the environment is changing, goals may also have to change to adapt to new circumstances. The goal lifecycle describes the process of formulating goals and monitoring the progress of plans. We also assume that we are operating with multiple agents. Thus, goals can be assigned and re-assigned to different agents.

 ActorSim's goal reasoning cycle is shown in Figure \ref{goal_cycle}. It consists of the following steps.
 
\begin{description}

\item[FORMULATE:] Determines which goals to create and adds them to the working memory.

\item[SELECT:] Of the possible FORMULATED goals, this step determines which goals to pursue, where the maximum number of selected goals depends on the available agents that can perform the action.

\item[EXPAND:] This step generates one or more plans to achieve the selected goal. More than one plan may be generated if there is more than one way of achieving the goal.

\item[COMMIT:] Chooses one expansion.

\item[DISPATCH:] A selected plan is dispatched to an agent and executed until completion or failure.

\item[EVALUATE:] Check events that impact the plan and determine if the goal is achieved or if there has been a plan failure.

\end{description}

Note that the diagram shows that there may be loops in the goal cycle. For example, if a plan fails, and there is more than one way of expanding a goal, the goal reasoning system may loop back to an alternative expansion. If a goal cannot be achieved in the present circumstances, it may defer the goal until conditions change and the goal becomes achievable. In the worst case, the goal may have to be dropped and new goals reformulated.

\begin{figure}[t]
\begin{center}
\includegraphics[width=\textwidth]{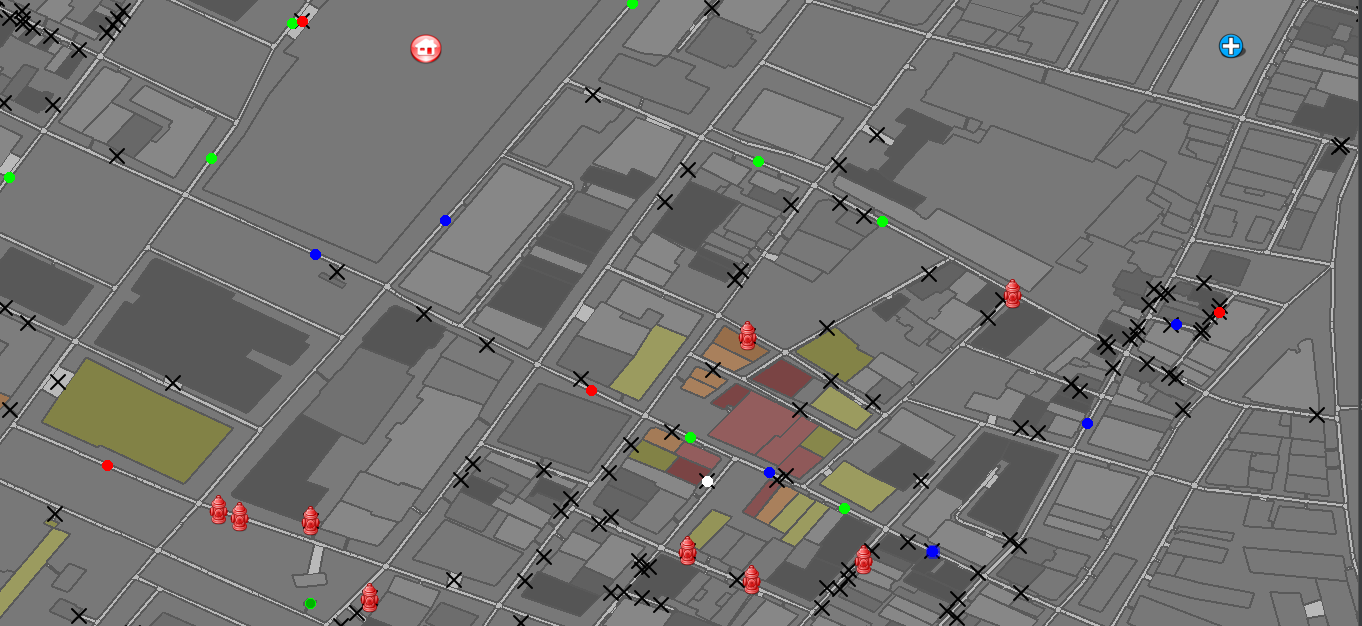}
\caption{A typical RoboCup Rescue Map. Civilians are shown as green dots, with darker hues indicating poorer health. A black dot means the victim has died. Fire trucks are red, ambulances, white and police yellow. Coloured buildings are on fire or in danger of collapse. Black 'X' denotes blockages in a road.} 
\label{typical_map}
\end{center}
\end{figure}

The main innovation presented here is that a human trainer monitors the progress of goal cycle and at steps one and two, where the system decides what goals to pursue and their priority, the trainer may intervene if they disagree with the system’s selection. The trainer can offer an alternative goal, at which point, the learning system updates its goal selection rules so that when a new situation arises that is similar to the present case, it will select a goal in line with the trainer’s choice. Rule construction is aided by comparing the present case with previous cases, in which the old rules were correct. The differences between the cases become the conditions in an exception rule to cover the new case.

The RDR system for goal reasoning has been tested in the domain of urban search and rescue, using the RoboCup Rescue Agent competition simulator \footnote{https://rescuesim.robocup.org}. This simulates a city environment after a disaster, such as an earthquake.  A typical scenario is shown in Figure \ref{typical_map}.

\section{Software Architecture}

\begin{figure}[b!]
\vskip 0.05in
\begin{center}
\includegraphics[width=5.5in]{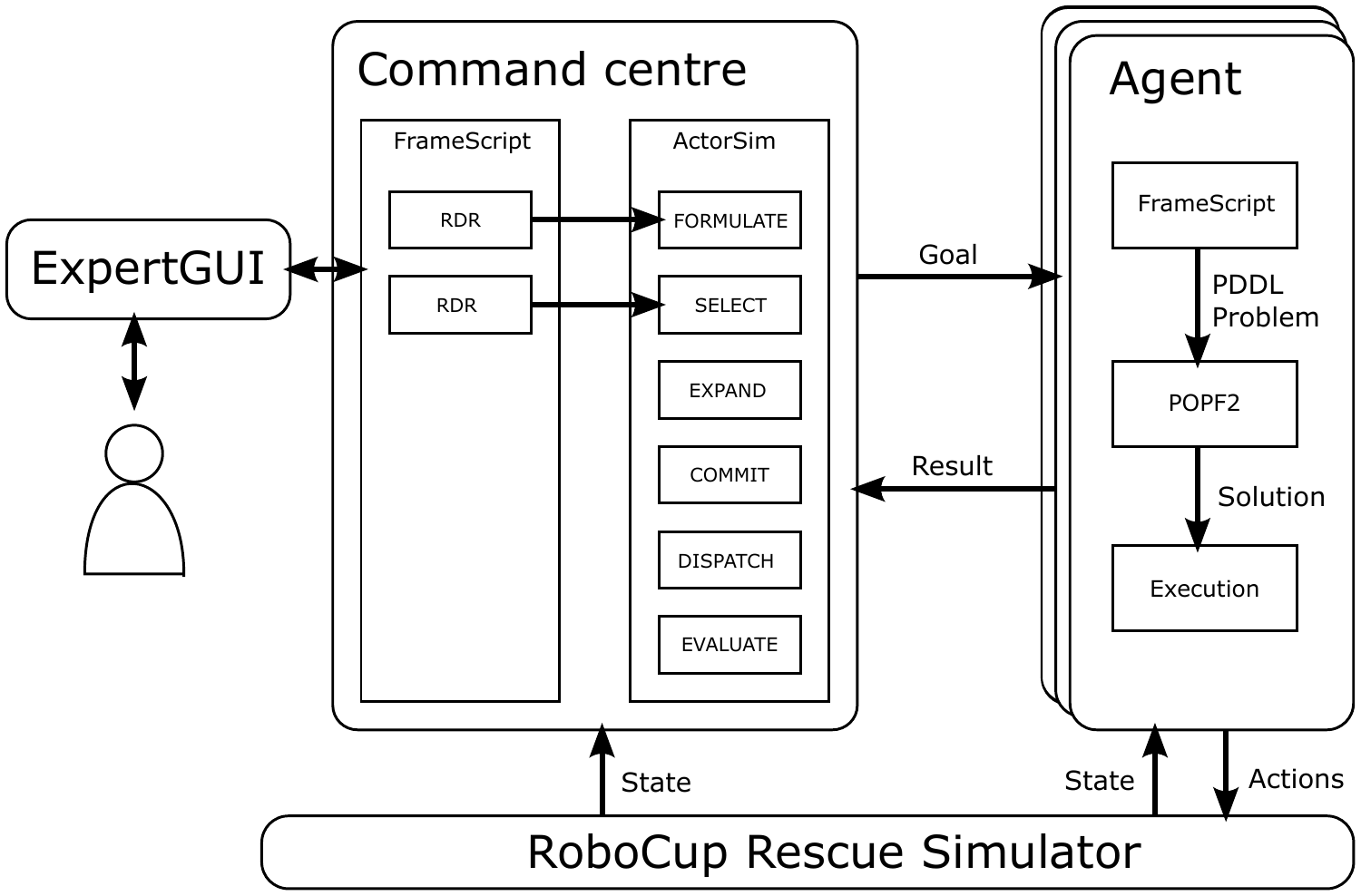}
\caption{Interaction between centralised goal planning and agent actions planning.}
\label{fig:planner-interactions}
\end{center}
\vskip -0.2in
\end{figure}

Figure \ref{fig:planner-interactions} shows the interaction between the main components involved in goal reasoning and plan generation.

\textbf{RoboCup Rescue Simulation Server:} The simulation server controls what happens in a city after a simulated earthquake. The RoboCup Rescue Agent Simulation competition originated from observations of the operation of emergency services following the Kobe earthquake of 1995. This caused city-wide damage and there was a recognition that much better planning was needed to coordinate emergency services to rescue survivors, put out fires and secure buildings. The simulator has a map of a city in which buildings can be unstable and in danger of collapse and they may be on fire. Civilians may be trapped in buildings, and possibly buried in rubble. Agents in the simulation include fire trucks, ambulances and police vehicles. Each year in the RoboCup competition a new city map is created, usually using the map of the host city.

\textbf{Team Agent:} At each time step in the simulation, the world is updated with movement of agents through the city, building fires spreading, building collapse damage, road blockages and damage suffered by human agents. A centralised command centre is in charge of assigning goals to \emph{platoon agents} (i.e. police, fire brigade and ambulance) which can perform actions in the simulation. In our case they are based on the Team Agent developed by NRL, using ActorSim. At each time step the platoon agents receive the information available to them through their sensors, which have a limited range. This information can be broadcast to other agents using a simulated radio channel. Based on these updates, the command centre assigns goals to the platoon agents, who invoke a partial order planner, via FrameScript rules, to formulate a sequence of actions that are sent back to the simulation server, one action per time step.

\textbf{Ripple-down rules:} As mentioned earlier, goal selection rules are learned by behavioural cloning. That is, the system learns by apprenticeship to a human trainer. ActorSim (combined with FrameScript) selects a goal for each agent. If the trainer is not satisfied with the selection, he or she can interrupt the simulation and suggest a different goal. At this point, the system will query the trainer by comparing the present situation with past cases in which the current RDR is correct. The difference between the cases enables the system to construct a new rule to add to the RDR. The trainer has the option to specify which differences are significant, thereby accelerating learning. There are two sets of RDR rules, the first is involved in the goal formulation, where the trainer can specify the conditions under which a new goal should be created. The second changes the priority of the selected goals.

\textbf{Expert GUI:} The graphical user interface that assists the trainer in controlling the simulator and creating the RDRs has several different views:

\begin{description}

\item[Map:] The user can see a partially observed state of the world in a map. The full state is filtered to show only information within the sensor range of the agents. Therefore the trainer can only make decisions with the information available to the agents.

\item[Goals:] A display shows the goals that have been formulated and assigned to an agent or are waiting to be assigned to an available agent.

\item[RDR:] An RDR panel shows the state of each agent, and is used to create and extend the ripple-down rules or goal formulation and prioritisation.

\end{description}

\subsection{Updating Ripple-Down Rules}

An RDR can be viewed as a kind of decision tree or nested if-then-else statement, with the addition of an exception clause. When an RDR gives a wrong conclusion, it may be because a rule is too general. In this case, it can be made more specific by adding an exception rule that excludes the current case from the old rule and adds a new conclusion to cover this case. If the RDR fails to produce an answer, it is too specific because there is no condition to cover it. Therefore the RDR must be generalised by adding a new rule as an else clause. Details of the RDR algorithm are given in~\citep{Compton:2021wk}.

Learning usually starts with a simple default rule called case $\emptyset$, e.g.

\begin{verbatim}
    if true then do_nothing
\end{verbatim}

\noindent
If this gives the wrong conclusion, the rule is specialised by adding an exception. For example, if a civilian or agent is buried, do\_nothing is an incorrect response, so an exception might state that if a person is buried then an \emph{unbury} goal should be placed in the goal queue.

\begin{verbatim}
    if true then do_nothing
        except if exists person(P) and buried(P) then unbury(P)
\end{verbatim}

\noindent
If a rule matches, but gives the wrong answer, it is generalised by adding an else clause. For example, if no one is buried, we may want an agent to explore, search for survivors.

\begin{verbatim}
    if true then do_nothing
        except if exists person(P)and buried(P) then unbury(P)
        else explore
\end{verbatim}

\noindent
As the name suggests, the main knowledge structure in FrameScript is a frame~\citep{roberts:1977uo}. Generic frames describes classes of objects, while instance frames describe particular objects that inherit properties from their parent generic frames. An example of a generic \emph{human} frame is shown below. 

\begin{verbatim}
human ako object with
    type:
        range [agent, civilian] 
    buriedness:
        range		
            [non_buried, buried]
    health:
        range [dead, critical, injured, healthy]
    goal:
        range		
            [none, unbury]
        if_needed	
            if true then none because case0
        if_replaced
            rdr_frame([buriedness])
\end{verbatim}

\noindent
The frame includes the slots, \emph{type}, \emph{buriedness}, \emph{health} and \emph{goal}, where the value of the slot is constrained to the set given in the range facet. The \emph{none} slot has an "if\_needed" daemon, which is evaluated when the value of the slot is requested. This invokes the RDR, in the default case, returning \emph{none}. The "because" clause is used to keep track of which cases caused an RDR update. The default "case0" is empty, but with each update a new \emph{cornerstone case} is added.

The default RDR evaluates the goal as "none" for all cases. The slot also has an "if\_replaced" daemon, which is called when the trainer says that the value returned by the RDR should be replaced by a new value. The \emph{rdr\_frame} function is predefined to perform the RDR update with the new value.

Now suppose in the simulation, there is a fire brigade agent that has been buried alive in debris inside a partially collapsed building. An instance frame for the brigade might be:

\begin{verbatim}
    frame(human_937073426, [human], [buriedness: buried]);
\end{verbatim}

\noindent
Querying the \emph{goal} slot for the frame above returns \emph{none}. Because the original RDR in the human frame evaluates to "none" in all cases: 

\begin{verbatim}
if_needed	
    if true then none because case0
\end{verbatim}

The trainer would indicate that this is the wrong conclusion, and the fire brigade agent should be unburied. That is, the previous conclusion should be replaced, invoking an RDR update. After the update, the "if\_needed" slot of the human frame  is:

\begin{verbatim}
if_needed	
    if true then none because case0
        except 
        if this buriedness == buried then
            unbury because case_brigade_210921154054
\end{verbatim}

An exception has been added to the RDR, such that is the "buriedness" slot is "buried", the "goal" is evaluated to "unbury". The cornerstone case, \emph{case\_brigade\_210921154054}, is a copy of the agent instance frame at that time the RDR was updated. A copy is required since the agent frame changes as the simulation continues. With the RDR above, all humans that are buried, will aim to be unburied.

\section{Operation of the System} 

\begin{figure}[t]
\vskip 0.05in
\begin{center}
\includegraphics[width=5.5in]{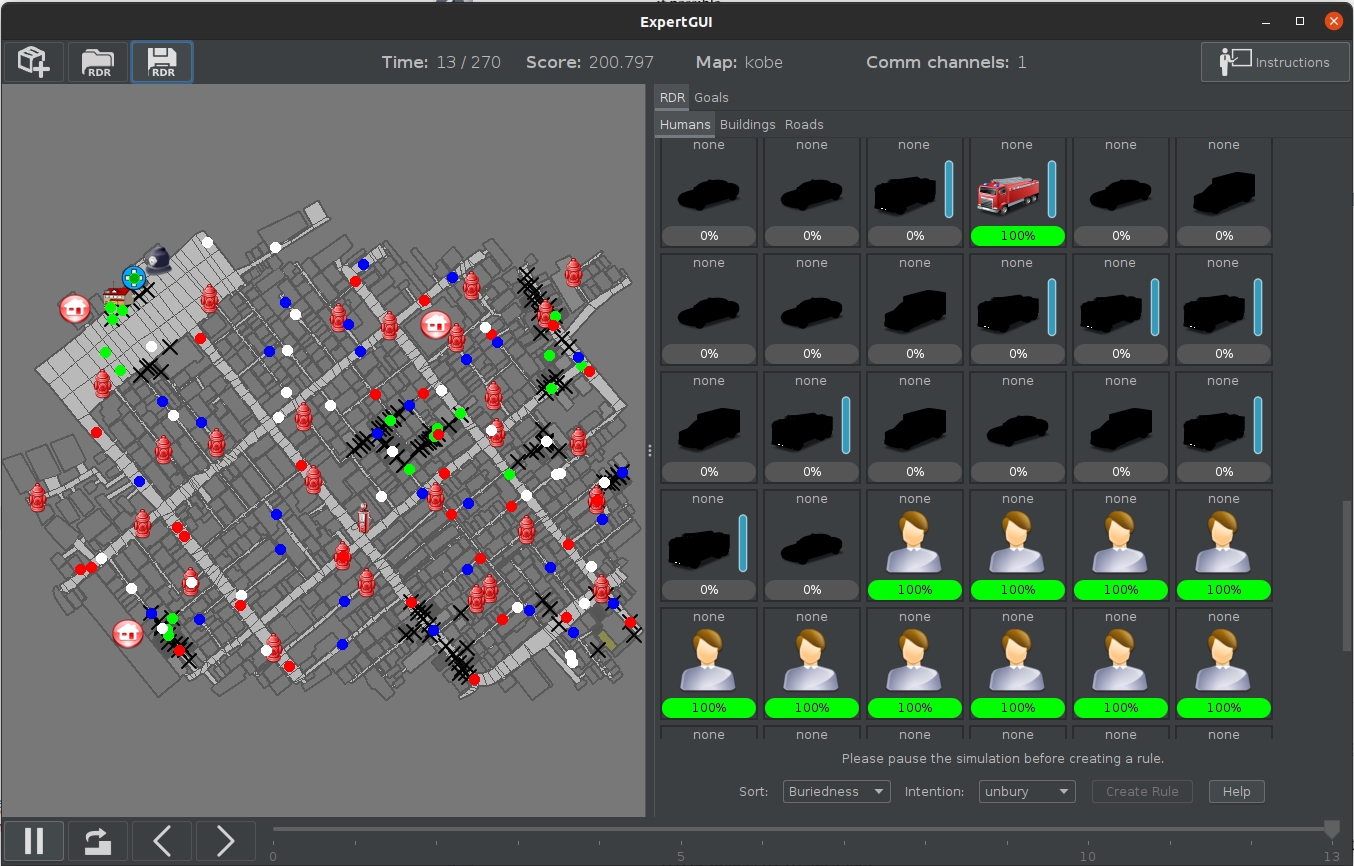}
\caption{ExpertGUI displaying the centre's knowledge of the world} 
\label{expert-gui}
\end{center}
\vskip -0.2in
\end{figure}

Figure \ref{expert-gui} shows a screenshot of the system with the city map on the left and the humans and their state on the right. The humans may be civilians or agents, including police, ambulance and fire trucks. The simulator also models the state of the buildings and roads. Buildings may be on fire or collapsed and roads may be blocked. Humans may be injured and buried.

\begin{figure}[t]
\vskip 0.05in
\begin{center}
\includegraphics[width=5.5in]{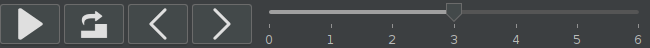}
\caption{Controls to start, pause the simulation and replay world states} 
\label{fig:timeline-panel}
\end{center}
\vskip -0.2in
\end{figure}

The simulation runs  with discrete time steps. At each step, the simulator notifies agents about what they can observe from their location, as well as the radio messages from other agents and central offices. Agents are required to determine an action to perform for the next time step and report back to the simulator.
The GUI displays information known to the centralised command centre, which aggregates information transferred over radio communication from all of the agents. The GUI provides buttons to pause and replay the simulation, to observe the state of the world at any time in the past (Figure \ref{fig:timeline-panel}).

FrameScript stores information observed or received by the centralised command centre, and as in the example above, also stores the RDRs. By iterating over all entities in the map and querying the goal slot, a list of goals is obtained from FrameScript. ActorSim then assigns goals from that list to agents and tracks each goal to completion.

\begin{figure}[t]
\vskip 0.05in
\begin{center}
\includegraphics[width=5.5in]{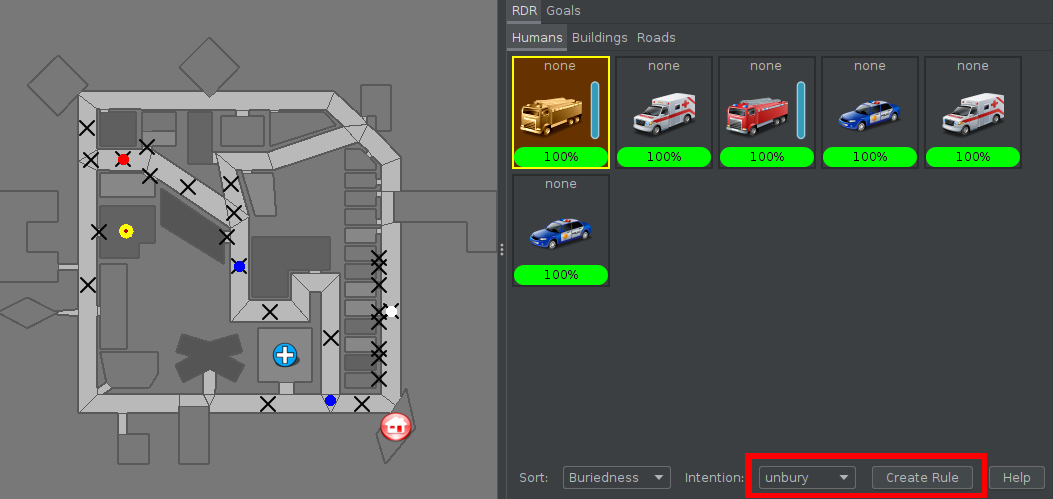}
\caption{An entity is selected for a rule to be created against} 
\label{entity-selection}
\end{center}
\vskip -0.2in
\end{figure}

\begin{figure}[t]
\vskip 0.05in
\begin{center}
\includegraphics[width=4.5in]{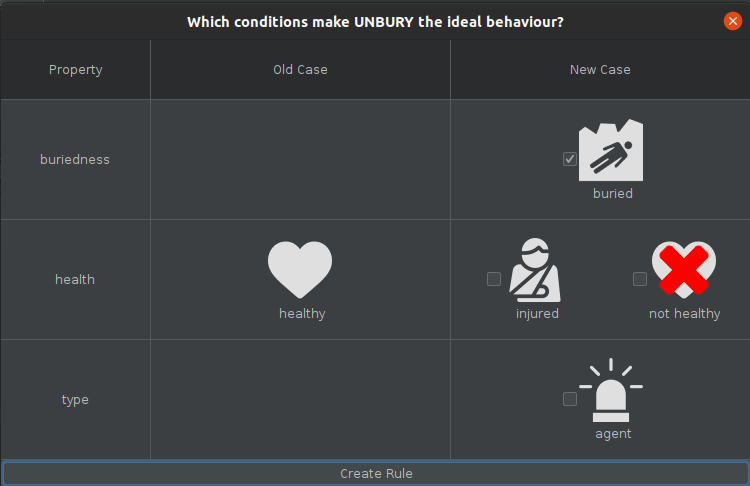}
\caption{RDRGUI allows visual interaction for creating RDRs}
\label{rdr-gui}
\end{center}
\vskip -0.2in
\end{figure}

To train the goal evaluation RDRs, the user pauses the simulator and selects the entity in the map for which a goal should have been determined (Figure \ref{entity-selection}), or was determined incorrectly, and assigns a new goal to the entity. The user must then select from a list, the properties of the entity that justify the goal's creation. Figure \ref{rdr-gui} shows how a new rule is added. Each time a new rule is added, the system keeps track of the cases that satisfy that rule.  When a new case arrives that causes the rule to fail, the system displays an old case that satisfies the rule and the new case. The trainer only needs to identify the difference between the cases. These differences are used to create the conditions for a new rule. Any entities that satisfy the conditions will be assigned the new goal in future time steps. 

For the example shown in Figure \ref{rdr-gui}, the trainer indicates that the relevant difference is $buriedness==buried$, then the initial RDR:

\begin{verbatim}
if true then none because case0
\end{verbatim}

is updated to this:

\begin{verbatim}
if true then none because case0
    except 
    if this buriedness == buried then unbury because case_brigade_1
\end{verbatim}

where the cornerstone case\_brigade\_1 is:

\begin{verbatim}
case_brigade_1:
    buriedness: buried
    health: injured
    type: agent
\end{verbatim}

Often there is a delay between the system making a mistake and the trainer recognising the mistake. To allow for this, the simulator can be rewound to the time when the error occurred so that the correct state of the world is used to obtain the properties of the entities.

A separate RDR is used to define the ordering of goals according to their importance. The goal ordering in ActorSim~\citep{roberts_goal_2021} implements a partial goal ordering. To change the order, the user selects two goals, and defines the ordering of the two. The type of goal is used in generating the ordering rules. ActorSim is then responsible for assigning the high priority goals (SELECT) by interrupting any lower priority goals (PREEMPT) that have already been assigned to agents.

Given that we only need to define a partial order between the goals, the RDR only needs to define the relation between two types of goals, for example, the RDR that gives precedence to rescue goals over scouting is defined as:

\begin{verbatim}
    if true then false
	    except if GoalA == rescueGoal and GoalB == scoutGoal 
	        then true because before(rescueGoal, scoutGoal)
\end{verbatim}

Once that the formulation and goal ordering rules have been applied, the goal life cycle dispatches the goals to the particular agents (ambulance, police, fire brigade).
On the agent's side, each agent has its own state of the world stored in FrameScript. The state representation simply mirrors the state information from the simulator. When the agent receives a goal from the central command centre, relevant information from FrameScript is extracted to produce a PDDL problem file.
This is passed to the POPF2 partial order planner~\citep{coles_forward-chaining_2010, coles_popf2_2011}, which returns the sequence of actions to be executed by the agent.

\section{Evaluation}
Table \ref{tab:test-cities} shows three different scenarios that were used to evaluate the system. The test city is a basic scenario provided by the rescue simulation for testing and development of the agents. Kobe and Montreal are part of the standard RoboCup competition, the first has a higher number of agents to control, while the second has a bigger map with a smaller amount of agents.

\begin{table}[h]
    \centering
    \begin{tabular}{|c|c|c|c|c|}
        \hline
         & Test city & Kobe city & Montreal city \\
        \hline
         Civilians & 5 & 200 & 100 \\
        \hline
         Agents & 3 & 90 & 36 \\
        \hline
         Buildings & 37 & 757 & 927 \\
        \hline
         Roads & 58 & 1602 & 3059 \\
        \hline
    \end{tabular}
    \caption{Number of entities for each testing scenario.}
    \label{tab:test-cities}
\end{table}

To demonstrate a simple update sequence, we used the ExpertGUI to run the simulator with the test city. In the simulation, \emph{fieryness} defines how advanced is the fire in a building, where \emph{destroyed} is the last stage. The trainer wants to douse all fires that have not consumed a building completely. Additionally, scouting the buildings is needed to gather information, so the following rules for buildings are created and saved:

\begin{verbatim}
if true then none because building0
    except
    if this scouted == no 
        then scout because building1
    else
    if this fieryness == heating 
        then douse because case_building_1
    else
    if this fieryness == burning 
        then douse because case_building_2
    else
    if this fieryness == inferno 
        then douse because case_building_3
\end{verbatim}

\noindent
The trainer also creates RDRs for goals associated with roads. In this scenario, an agent is stuck in a blocked road and needs it to be cleared. Since the scenario is set immediately after an earthquake, the trainer only wants to unblock roads that have been requested by an agent, leaving less urgent roads aside.

\begin{verbatim}
if true then none because road0
    except
    if this requested == yes and this blocked == yes 
        then unblock because case_road_1
\end{verbatim}

\noindent
 The keyword \emph{this} means that the expression is referring to a slot in the same frame as the RDR (similar to \emph{self} in Python).

\noindent
As in the previous examples, the trainer also constructs an RDR for humans:

\begin{verbatim}
if true then none because case0
    except
    if this buriedness == buried 
        then unbury because case_brigade_1
\end{verbatim}
Any human, civilian or agent, must be unburied.

All agents can \emph{scout}, but \emph{rescue}, \emph{clear} and \emph{douse} can only be assigned to ambulances, police and fire trucks, respectively. Therefore, the trainer defines a partial order where rescue, clear and douse goals have precedence over scouting. The following RDR returns \emph{true} if the first goal argument should be performed before the second goal argument.

\begin{verbatim}
if true then false
	except 
	if GoalA == rescueGoal and GoalB == scoutGoal 
	    then true because before(rescueGoal, scoutGoal)
	else 
	if GoalA == clearGoal and GoalB == scoutGoal 
	    then true because before(clearGoal, scoutGoal)
	else 
	if GoalA == douseGoal and GoalB == scoutGoal 
	    then true because before(douseGoal, scoutGoal);
\end{verbatim}

Frames, including RDRs can be saved and imported into new scenarios. For example, the rules above were created for the simple test city but can then be imported for the Kobe map. Training can then continue for this scenario. The trainer updates a new case for the roads where a road is unblocked if a civilian is trapped in it. The new rules for roads are:

\begin{verbatim}
if true then none because road0
    except
    if this requested == yes and this blocked == yes 
        then unblock because case_road_1
    else
    if this has_civilians == yes and this blocked == yes 
        then unblock because case_road_2
\end{verbatim}

Lastly, to validate that the system has learned the combined rules of the previous two maps, the system was tested on the Montreal map. This time there was no need to perform any updates as the rules learned for the previous maps also worked for the Montreal map.

\begin{table}[h]
    \centering
    \begin{tabular}{|c|c|c|c|c|}
        \hline
         & Test city & Kobe city & Montreal city \\
        \hline
         Goals & 99 & 1305 & 1380 \\
        \hline
         Time steps & 300 & 270 & 270 \\
        \hline
    \end{tabular}
    \caption{Number of goals created and duration of the simulation.}
    \label{tab:goals}
\end{table}

In our evaluation the system was able to successfully assign goals to the agents and execute the commands in the simulation. However, we have not yet attempted to optimise the code. In the RoboCup competition, agents are required to reply with their commands within one second, otherwise the server skips their turn. Our system can take from one to ten seconds to respond depending on the number of entities in the map and the number of goals. This is because we naively iterate through all the entities, evaluating RDRs. The performance can be improved substantially by using a smarter evaluation strategy. Another aspect that has been simplified in our system is the reliability of the radio channels. In the competition, messages between agents can be dropped or corrupted, affecting the information known to the agents and goal assignments.

 \section{Conclusion}

The original ActorSim goal reasoning framework has been augmented by the addition of a scripting language (FrameScript) that incorporates a knowledge acquisition mechanism. The system is capable of assigning goals to multiple agents and the invoking a partial order planner to achieve those goals. Development continues to shift code for strategies from Java to FrameScript to make it easier to adapt the system to different domains.

A significant extension to be considered for future work is the ability to model hierarchical command and control structures. Thus, at the top level, ActorSim may allocated a complex task to an agent, but that agent may, itself, consist of subordinate agents. The task may decompose into subtasks that must be allocated to this subordinates, so we could invoke ActorSim recursively down the command and control hierarchy.

The motivation for using RDRs to represent goal selection strategies is that it will enable the extraction of explanations from the rules. Moreover, because rule creation is triggered by particular cases, it will be possible to give explanations in terms of examples, that are much easier for humans to understand. A further advantage of FrameScript is that it includes a dialogue management system that can facilitate question answering.

\begin{acknowledgements} 
\noindent
This work was funded by the Australian Defence Science and Technology Group (DSTG), in collaboration with the U.S. Naval Research Laboratory (NRL), as part of a joint program supported by the U.S. Air Force Office of Scientific Research (AFOSR). We thank Martin Oxenham for his support through the DSTG and Doug Riecken for his support through AFOSR.
\end{acknowledgements} 

\vspace{-0.25in}

{\parindent -10pt\leftskip 10pt\noindent
\bibliographystyle{cogsysapa}
\bibliography{rdr_gr}

}


\end{document}